# A Spatial-temporal Graph Deep Learning Model for Urban Flood Nowcasting Leveraging Heterogeneous Community Features

Hamed Farahmand[1,*], Yuanchang Xu[2], Ali Mostafavi[1]

[1]Zachry Department of Civil and Environmental Engineering, Texas A&M University, College Station, TX

[2]Department of Computer Science and Computer Engineering, Texas A&M University, College Station, TX

**Abstract**

The objective of this study is to develop and test a novel structured deep-learning modeling framework for urban flood nowcasting by integrating physics-based and human-sensed features. We present a new computational modeling framework including an attention-based spatial-temporal graph convolution network (ASTGCN) model and different streams of data that are collected in real-time, preprocessed, and fed into the model to consider spatial and temporal information and dependencies that improve flood nowcasting. The novelty of the computational modeling framework is threefold; first, the model is capable of considering spatial and temporal dependencies in inundation propagation thanks to the spatial and temporal graph convolutional modules; second, it enables capturing the influence of heterogeneous temporal data streams that can signal flooding status, including physics-based features such as rainfall intensity and water elevation, and human-sensed data such as flood reports and fluctuations of human activity. Third, its attention mechanism enables the model to direct its focus on the most influential features that vary dynamically. We show the application of the modeling framework in the context of Harris Count, Texas, as the case study and Hurricane Harvey as the flood event. Results indicate that the model provides superior performance for the nowcasting of urban flood inundation at the census tract level, with a precision of 0.808 and a recall of 0.891, which shows the model performs better compared with some other novel models. Moreover, ASTGCN model performance improves when heterogeneous dynamic features are added into the model that solely relies on physics-based features, which demonstrates the promise of using heterogenous human-sensed data for flood nowcasting,

## Introduction

### Background

Flood nowcasting is a process by which areas at imminent risk of inundation can be identified using the spatial and temporal features that convey information regarding current flooding status. As extreme weather events accompanied by heavy precipitation occur more frequently, causing catastrophic flood events, flood nowcasting has become an essential capability for communities to better respond to the impacts of these events [1]. Flood nowcasting enables predictive flood monitoring, the ability to anticipate imminent flood risks and impacts and situational awareness as an extreme weather event unfolds [2]. Departing from the standard flood monitoring approaches using hydraulic and hydrological (H&H) models that predict flood inundation levels for hazard mitigation and infrastructure improvements prior to flood events [3], [4], flood nowcasting focuses on near-future prediction (e.g., next few hours) of spatial and temporal flood status based on the current status of flooding. The traditional approaches for flood monitoring [5] do not provide certain essential information (e.g., what areas will be inundated within the next few hours). Nowcasting will enable public officials, emergency managers, responders, and residents to better tailor decisions and actions by enhancing situational awareness during response and recovery [6]. Hence, urban flood nowcasting facilitates identifying areas that will require emergency aid in hours immediately ahead and areas that need issuance of evacuation notices due to the high risk of flood inundation. This forewarning is critical for reducing the adverse impacts of flood events. It also facilitates taking proper managerial actions to control flood inundation using hydrological infrastructures, such as flood gates and pumps [7]–[10]. The main approach for sensing flood status is the use of rainfall and stream gauges; however, due to cost and maintenance limitations, the number of these physical sensors is limited, which affects proper observability of flood status [11], and hence, flood nowcasting. New techniques for enhancing situation awareness and emergency response actions leverage heterogeneous community-scale datasets (including both physical sensors and crowdsourced data) in advance to provide the predictive capability to infer the flooding status for the near future in spatial units (e.g., zip code, census tract, and neighborhood), information [12]–[14].

Multiple studies have been conducted to develop predictive tools using a wide range of physics-based features and quantitative techniques. Conventionally, H&H simulation models are used for predicting the extent of flooding in urban areas using geomorphological hydrodynamic features to estimate water depth in urban areas [15], [16]. These models often rely on the data collected from rainfall and flood gauges to provide an estimate of the spatial extent of flood propagation [6], [17]. Despite their satisfactory accuracy and predictive performance, extensive computational cost and the sparsity of the reliable hydrological data in urban areas limit the existing physics-based H&H models [57], [58][18] for providing near-future estimation for spatial-temporal propagation [19]. To complement the standard models, recent studies tested data-driven models based on harnessing data sources, such as satellite images, crowdsourced data, and remote-sensing data, that can help estimate flood status in near future timeframes [3], [20]–[23]. Also, a growing number of researchers have used the predictive capability of various machine learning (ML) models for flood predictive monitoring [24]–[29]. These models can include more community features than tradition models to forecast flood status, which facilitates capturing the large number of heterogeneous community features needed for flood nowcasting [10], [30].

In the following sections, we review the state-of-the-art in application of deep-learning models for flood nowcasting to identify gaps in the existing literature. We focus particularly on two major gaps: 1) the absence of a model architecture that enables capturing spatial and temporal dependencies in flood propagation and dynamically identifying influential features, and 2) limited efforts for integrating human sensing as an approach for collecting and extracting valuable temporal and spatial data. We also review the use of heterogenous human-sensed data as a supplement for flood nowcasting to show the gap in the knowledge regarding the proper use of such data for improving the urban flood nowcasting models. Accordingly, we present and test a novel graph-based deep-learning models that enable capturing spatial dependencies, as well as heterogeneous human-sensed features in flood propagation. We demonstrate the application of the proposed model in the context of the 2017 Hurricane Harvey flooding in Harris County, Texas.

### Related Works
#### Deep Learning for Flood Nowcasting

Advances in machine learning techniques are responsible for the emergence of deep learning (DL), a sub-domain of ML that employs deep artificial neural network architectures and gradient descent algorithms for yielding more robust and computationally efficient predictive models [31]–[33]. Deep neural networks have been increasingly used for tasks that support flood predictive monitoring, such as flood depth mapping and flood detection. Multiple studies have applied DL techniques to improve the predictive performance of physics-based flood nowcasting models. For example, a convolutional neural networks (CNN) have been used in combination with conditional generative adversarial networks (cGAN) to improve the performance of physics-based flood forecast models [37]. In addition, a combined empirical mode decomposition (EMD) algorithm and encoder-decoder long short-term memory (En-De-LSTM) architecture have proved to yield a better prediction of peak flow values of streams during floods [38]. Recent data-driven models rely purely on the capability of DL models for flood prediction. For example, streamflow prediction using an integration of stacked autoencoders (SAE) and back propagation neural networks (BPNN) show higher accuracy compared with other tested ML models [39]. Also, Gated Recurrent Units (GRU)-based network architecture has been utilized for predicting the time series of stream sensors used for flood monitoring [40]. In a recent work by Dong et al. [37], a Fast GRNN-FCN (fast, accurate, stable, and tiny gated recurrent neural network-fully convolutional network) was proposed for forecasting the water level in channel network sensors to provide flood signals in flood control network [41]. While the use of DL models for flood prediction is becoming prevalent in the literature and practice, the current research trends lack a computational data-driven modeling framework that enables a near-future prediction of flood status in spatial blocks (e.g., census tracts or zip codes). This gap is due mainly to: (1) inability of the existing models to capture the spatial interdependencies; (2) limitations in extracting features that provide indication of flood status in spatial blocks (due mainly to a limited number of physical sensors). The inability to predict near-future flood status in spatial blocks is a major hindrance to flood nowcasting. To address this gap, in this study, we propose a spatial-temporal graph deep-learning model.

Incorporating attention mechanisms into spatial-temporal deep-learning models for flood prediction elicits superior results compared to other state-of-the-art model architecture, and also improves interpretability of the model results [42]. These studies often leverage the ability of different DL models for time-series forecasting and early warning detection utilizing sensors that collect rainfall and streamflow data. Nevertheless, most recent studies have: (1) employed DL architectures that enable incorporating spatial correlation, and (2) created DL architectures that enable more feature incorporation. For spatial correlation, graph neural networks can capture the spatial similarity of model units [43] while an attention mechanism that enables the model to focus on the characteristic data when processing large numbers of features [44] and enable use of heterogeneous data to provide reliable prediction in urban units. In the next sections, we discuss the application of graph neural networks for spatial and temporal prediction as well as using heterogeneous data in flood predictive monitoring, which form the points of departure for this study.

   Graph Neural Networks for Spatial-Temporal Prediction

Graph neural networks generalize convolution to data in a graph structure [45]. With their superior capability to characterize spatial and temporal dependencies for time-series predictions, GCNs characterize networked data with spatial and temporal dependencies for time-series prediction using spatial and temporal convolutions. These models (referred to as spatio-temporal graph convolutional network (STGCN) models) are used for prediction problems such as traffic flow prediction [46], [47], disease diagnosis [48], bike-demand prediction [49], point-of-interest (POI) recommendation [47], pedestrian flow prediction [50], trajectory prediction [51], and road network flood inundation prediction [2]. STGCN model architectures have been developed based on the problem characteristics. For example, dual-channel based graph convolutional networks (DC-STGCN) consider both daily and weekly correlation of the traffic data [52]. Discriminative spatio-temporal graph convolutional network (DSTGCN) were used for action recognition in to inner-class action distribution [53]. Wang et al. (2018) developed an auto-STGCN algorithm that facilitates the detection of the optimal STGCNs models automatically using a reinforcement learning technique [54]. An attention mechanism allows DL models to focus more on the useful parts of features [44]. In graph neural networks, the attention mechanism allows the model to learn a dynamic and adaptive combination of the adjacency matrices and select the most relevant information [55]. Attention-based GCNs adaptively capture dynamic spatial and temporal correlation of heterogeneous data and its interpretability power [56]. The combination of attention mechanism and STGCN structure, therefore, could provide a powerful testbed for problems in which heterogeneous features with complex spatial and temporal correlation exist. The application of attention-based STGCNs in the literature, however, is limited to traffic flow prediction [56]. Because of the characteristics of the urban flood nowcasting problem, attention-based STGCNs may provide models that could account for spatial interdependencies, as well as for the temporal correlations among features related to flood inundation status.

Heterogeneous Human-sensed Data for Flood Nowcasting

To complement the information sensed by physical sensors, other sources of data with distinct levels of reliability, aggregation, and the need for preprocessing have been tested in recent studies [59]. Satellite images, drone-recorded videos and images, and images captured by other cameras provide reliable information; however, limitations of data acquisition and challenges in data processing restrict extensive use of such data for flood predictive monitoring and vulnerability assessment [60]–[63]. Blumberg et al. (2015) [1] employed hurricane-related photos provided by volunteers to simulate flood inundation during Hurricane Sandy in Hoboken and Jersey City, New Jersey. Yin and Wilby (2016) used the crowdsourced data to validate their simulated flood scenarios and the associated impacts on land subsidence in Shangha, China. On the other hand, human-sensed crowdsourced data have become more available in different formats that can provide geo-located information regarding the flood status in a timely manner. For example, studies have analyzed anonymized social media content using ML and DL techniques and employed the extracted information for enhancing flood situational awareness [64]–[67]. In another study example, Huang et al. (2018) [68] integrated tweet data gathered by remote sensing and river water gauges to improve near real-time flood inundation maps. Moreover, the tweet activity data has also proven to expedite the detection of flood inundation and flood-related events when combined with satellite flood signals [69]. However, there are limitations in terms of content analysis and ensuring the credibility of the extracted information from social media [70]. Furthermore, social media data might be biased by factors such as distance to impacted areas, the popularity of the user, and demographic characteristics of users [71]. Recently, the digital trace of human activities (such as cellphone and location-based data) has also been deployed for flood prediction. The rationale is that the changes in the level of human activity and the concentration of human activity can provide signals regarding flood status [72]. The combined use of different sources of data—physical flood sensors data, crowdsourced social media data, and telemetry-based human activity data provides opportunities to gather a more extensive set of indicators related to flood status for use in flood predictive monitoring [73]. Integrating such heterogeneous data requires a modeling framework that is able to recognize and focus on key data features. The attention-based STGCN model proposed in this study enables leveraging heterogeneous datasets to capture features related to flood inundation status for flood nowcasting.

### Point of Departure

The review of the current state of the art shows two gaps in the knowledge for urban flood nowcasting: (1) the absence of a deep-learning structure that combines attention mechanism and graph-based convolutional network structure for extracting information from heterogeneous features with complex spatial and temporal correlation; and (2) the lack of a proper flood nowcasting modeling framework for integrating heterogeneous human-sensed features that can carry valuable flood-related information along with the physical sensor data. Recognizing these gaps, this study presents a deep-learning modeling framework including an attention-based spatial-temporal graph convolution network (ASTGCN) model and streams of data that could be collected as a flood event unfolds, preprocessed, and fed into the prediction model to consider spatial and temporal and dependencies and enable reliable urban flood nowcasting. The proposed model was tested in the context of flooding caused by the 2017 Hurricane Harvey in Harris County, Texas. The model performance and its implications for flood nowcasting, as well as enhancing situation awareness, are discussed. The novelty of this study is the creation an attention-based deep learning framework that addresses major limitations in the application of data-driven techniques for flood nowcasting by (1) focusing on graph-based architectures that enable co-location dependency between urban units for considering the spatial aspect of flood propagation, (2) identifying and processing various heterogeneous physics-based and human-sensed data that carry information for inferring flood status in spatial units, and (3) utilizing an attention-based time-series forecasting architecture for considering the temporal aspect of flood prediction and focusing on information with higher importance when processing large amounts of heterogeneous features.

## Methods

### Problem Definition and Abstraction

In this study, we model the study area as a network of census tracts to capture the spatial interdependence in urban flood propagation and recession. We used the census tract as the spatial unit since its scale is neither so coarse as to lose the resolution nor so fine as to lose observability of flood status due to missing data. This moderate status makes the census tract a suitable spatial scale for aggregating and interpolating both human-sensed data and physics-based data while maintaining data accuracy and keeping it informative for flood nowcasting. We created an undirected graph $G = (V, E, A)$, where $V$ is the set of $N$ nodes, each representing a census tract in the study area; set $E$ includes edges in graph $G$ that represent the connection between different nodes; and matrix $A_{N \times N}$ is the adjacency matrix of graph $G$. Entries of matrix $A$ are determined based on the proximity and the extent to which two census tracts have similar features that potentially influence their flooding status. Therefore, matrix $A$ is built upon the distance between census tracts and a set of static features, such as elevation, land use, and distance to stream, that impact the flooding status of particular areas [74]. At each timestep, each node in the graph $G$ holds a vector of temporal features (more discussion about the features are provided in the next section) that contain information that is used as the model input for nowcasting flood in the model. These temporal features capture various physics-based and human-sensed data inputs that are aggregated and preprocessed into the same sampling frequency. Figure 1 shows a schematic representation of the graph model, as well as static and dynamic features that are used for feeding the model for flood nowcasting.

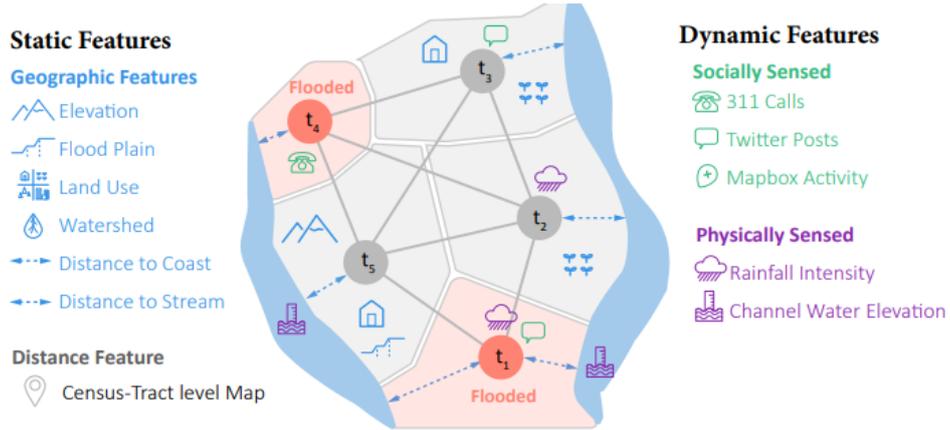

*Figure 1. Schematic representation of the problem abstraction; the study area is modeled as a graph; static features and distance are used to determine weights, and physics-based and human-sensed dynamic features are used for predicting the extent of flooding.*

Overview of the Model Development and Evaluation

Figure 2 shows the overview of the steps for the development and evaluation of the model. Overall, implementation and assessment of the performance of the framework involves four steps: data collection, data preprocessing, model development, and model evaluation. First, we present the data used for the development of the model. The data includes ground truth data: static features, which represent the dependency between flooding status of different areas in the adjacency matrix; and dynamic features that provide indications of temporal propagation and recession of urban flooding in each census tract. We also elaborate on the data preprocessing needed for the preparation of static features and the construction of time series of the dynamic features. Then we present the model architecture and mechanisms used in the DL model for urban flood nowcasting. Finally, we discuss the performance evaluation metrics of the model, parameter tuning for optimizing the model performance, and comparison of the model performance with other state-of-the-art models.

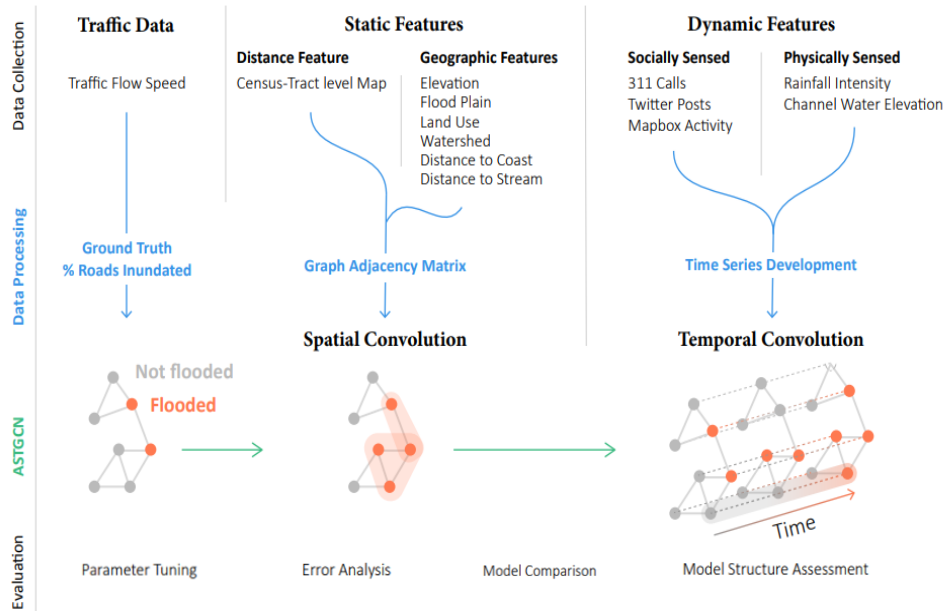

*Figure 2. Overview of the model framework, including steps for collecting and preprocessing ground truth and features, developing the ASTGCN model architecture, and evaluating the performance of the model and model performance.*

## Data Collection and Preprocessing

### Ground Truth

We used traffic condition data for 19,712 road segments in Harris County provided by the private company INRIX as proxies to determine if a certain road section was flooded. INRIX collects location-based data from both sensors and vehicles. The INRIX traffic data contains the average traffic speed of each road segment at 5-minute intervals and their corresponding historical average traffic speed. Each road segment's identification information, such as name, geographic locations defining its head and end coordinates, and length, is also available from the INRIX data set. Road segments flooded due to Hurricane Harvey can be identified by detecting the road segments with NULL values for their average traffic speed [19]. We filtered the road segments with $speed\ limit \geq 30$ to only account for the main roads in inundation estimation. We found that this filtering helps to reduce the data imbalance problem by capturing more flooded segments. The filtered data is used to determine the percentage of roads flooded as the indicator of the flood extent in census tracts. To do so, we characterized the flood status of each census tract as the ratio of flooded roads to the total number of roads.

### Static Features

Static features were used to develop the adjacency matrix and assign weights of connection between nodes in the graph model. We developed the adjacency matrix primarily based on the distance between the centroids of census tracts. In addition, we incorporated the impact of six static features that characterize flood propagation in an area in the adjacency matrix. The rationale is that nodes that have similar static features would have similar flood propagation behavior. Table 1 shows static features and the description of how they are calculated. These features were collected for each census tract. For the features available for each point, the value of the centroid of the census tract is considered. Elevation from the sea level was calculated using the digital elevation model (DEM) of the study area. Distance to Galveston coast and distance to closest main streamflow were calculated by mapping the study area and the streams that discharge stormwater from the area into Galveston Bay. Moreover, we coded 22 watersheds within the study area, and each census tract was associated with the watershed within which its centroid falls. Similarly, we mapped the 100-year floodplain and determined whether the centroid of the census tract falls inside the floodplain. The resulting binary variable was then used as a static feature. Finally, we used the land-use map of Harris County and determined the ratio of residential area to total land area as a feature that is a determinant of the land properties.

*Table 1. Static and dynamic features used for urban flood nowcasting*

| Influencing factor | Feature |
| --- | --- |
| **Static Feature** | |
| Floodplain | Whether or not the area is inside the 100-year floodplain |
| Land use | Percentage of the residential area |
| Watershed | The watershed that the area falls inside it |
| Distance to coast | Distance to Galveston coast |
| Distance to stream | Distance to the closest main stream flow |
| **Dynamic Feature** | |
| **Physics-based features** | |
| Short-term rainfall intensity | Estimated accumulated rainfall in past 2 hours |
| Long-term rainfall intensity | Estimated accumulated rainfall in past 24 hours |
| Water elevation | Estimated ratio of water level to the flooding threshold, based on average readings of two closest channels |
| **Human-sensed features** | |
| Flood reports | Number of reported flooding in the neighborhood through 3-1-1 platform |
| Social media activity | Number of flood-related filtered tweets |
| Human activity | Activity index of telemetry-based digital trace of human activity |

Dynamic Features

Dynamic features capture temporal changes that can indicate the flood propagation and can be used by the model for flood nowcasting. We considered both physics-based and human-sensed features. For physics-based features, we used the data recorded by the 175 flood gauge stations in Harris County. These food gauge stations are located on the main channels and bayous to provide residents with timely information on rainfall accumulation and water elevation in the stream [75]. We collected the rainfall and stream elevation from the official website of Harris County Flood Control District [75]. We constructed three time series for each census tract based on the flood gauge data, including short-term rainfall intensity, long-term flood intensity, and water elevation. For short-term rainfall intensity, we used the accumulated rainfall in the past 2 hours recorded by the flood gauge. For long-term rainfall intensity, we used the accumulated rainfall in the past 24 hours recorded by the flood gauges. Also, we used the ratio of recorded water elevation to the threshold elevation of flooding in each flood gauge as the water elevation indicator. It should be noted that the frequency of readings of rainfall and water elevation varies across time; in such cases, we performed interpolation and extrapolation to extract the value of the time series based on the available readings. The number of flood gauges is fewer than the number of census tracts; therefore, we used the weighted average of readings of the two closest flood gauges to determine measurements for each census tract. Weights are proportional to the inverse of the distance between the centroid of the census tract and the flood gauge. Figure 3 illustrates the process for determining physics-based features for each census tract based on the flood gauge data.

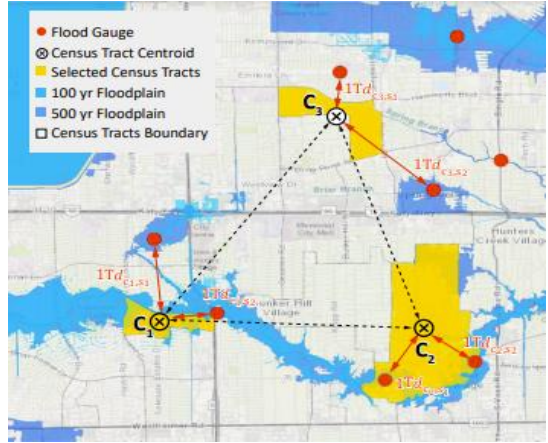

*Figure 3. Schematic illustration of calculating census tracts' distances and weights of two closest flood gauges as inputs to the census tract physic-based features.*

Physics-based features provide a reliable source for indicators needed for flood nowcasting; however, due to limitations such as sparsity of data points and lack of sufficient data (limited number of physical sensors) for inferring flood status in near future, we used a number of human-sensed data types to supplement the data needed for flood nowcasting. We used three different types of human-sensed data: records of 3-1-1 flood reports, Twitter activity, and the telemetry-based digital trace of human activity. We collected 4,275 flood-related 3-1-1 reports for the study period from the official website of the City of Houston [76]. Then, we filtered reports based on report type so that only reports that indicated flooding were included. More 3-1-1 flood reports during a certain timeframe in an area indicate a higher risk of flooding [74]; thus, we spatially aggregated the number of reports in each timestep and created a time series showing the number of floods reported through the 3-1-1 platform for each census tract. Social media platforms are another means by which people disseminate information regarding flooding in near real-time. Hence, the relevant data collected from social media can improve flood nowcasting. We incorporated flood-related information posted by Twitter users as an input for our flood nowcasting model. The geotagging feature of Twitter links tweets with accurate longitude and latitude of the location from which tweets originate [77]. Although a small percentage of tweets have geotagged, this small percentage generates thousands of tweets that provide reliable insights into flood status, especially areas lacking physical sensors. To examine social media attention, we collected tweets for the study time period (August 25, 2017, to September 2, 2017) in 84 super-neighborhoods in Houston. Twitter PowerTrack API (application programming interface) was used for collecting the 29,256 geotagged tweets during the study time period. Two filters were applied to ensure the relevance of the tweets. The first filter identifies the tweets, whose geotags like in our predefined bounding boxes, posted by the users whose profiles show their location Harris County. The second filter was the keywords (i.e., the names and abbreviations of the areas) that identify the tweets specifically related to the study area.

In addition to flood reports and social media activities, recent studies show that human activity fluctuations can signal flood inundation or other disaster-related impacts [78], [79]. To incorporate information regarding human activity in our flood nowcasting model, we obtained digital traces of human activities for the study timeframe from Mapbox. We chose Mapbox as the source of the telemetry data due to its ability to collect temporal and spatial telemetry-based human activity with a proper level of aggregation. Human activity is collected, aggregated, and normalized by Mapbox based on the geography information updates of locations of users' devices (such as cell phones) from applications that use Mapbox Software Development Kit (SDK). Human activity here refers to the density of digital traces recorded from user devices drawn from users of Mapbox SDK globally contributing to live location updates. (The data is gathered from app developers who access Mapbox data through the SDK. Mapbox records locations of users of the maps service.) Mapbox provided a 4-hour temporal resolution as raw data. In terms of spatial resolution, tiles represent square geographic areas approximately 100 meters per side, a size which varies depending on latitude. The more users located in a tile at time $t$, the greater the human activity index. Data might not exist for all spatial units, as data is derived from cell phone activity depending on the updates of the geography information of cell phone users. Moreover, to preserve privacy and the data aggregation process, traces are excluded from tiles with small numbers of users. The raw index of human activity is normalized. Normalization is compartmented separately by month and type of trace and yields a normalized activity index for each tile in each 4-hour time period of human activity provided by Mapbox. The normalized values range between 0 and 1. We created time series of human activity by aggregating tiles into census tracts and averaging the activity indexes for all the tiles that fall into a census tract in a certain timestep. Thus, we used linear interpolation to aggregate indexes of human activity for each 30-minute timestep as the time period considered for our model. Table 1 also provides a summary of dynamic features used for flood nowcasting in this study.

### ASTGCN Model

#### Graph Adjacency Matrix

The adjacency matrix in our model captures the co-location of census tracts and their dependency in terms of the state of flooding in terms of similarity of static features. Considering that the graph represents an area, and each node represents a census tract, co-location of two census tracts can imply similarities between their state of flooding. Therefore, we considered the distance between census tracts as the major determinant of the weights in the adjacency matrix. In addition to physical distance, we considered static features that imply similarity in flooding status of two areas. In particular, we considered features that influence flooding status in a flood-prone urban area: (1) whether the area is inside the 100-year floodplain, (2) distance to the closest main streamflow, (3) distance to the outlet (Galveston Bay in our study area), (4) the watershed in which that the area is located, and (5) the land-use pattern. To include these static features in our adjacency matrix, we created a vector of size five for each census tract containing the static features and calculated the Euclidean distance similarity for each pair of census tracts. To combine the impact of static features and co-location dependency, we used the weighted average of the Euclidean distance similarity and the physical distance. Based on the early experiments on the model for tuning the weights for the adjacent matrix, we found that choosing 0.1 as the weight for Euclidean distance similarity and 0.9 as the weight for physical distance yields the best result.

#### Model Architecture

We adopted the ASTGCN model architecture design from the model proposed by Guo et al. (2019) [80] that was developed primarily as an attention-based graph convolutional network for forecasting traffic flow. The original model framework includes three independent input components and employs information fusion to consider different temporal properties of the traffic flow and to deal with the seasonality of the traffic data. In the case of flood nowcasting, however, there is often no seasonality in the temporal changes of major features—such as rainfall, stream elevation, and human activity—during the hazard period. Hence, we used a single input component in our architecture that consists of time series of three physics-based and three human-sensed dynamic features recorded for each node of the graph. Thus, given the six dynamic features, and $N$ nodes in the graph model of the area, all the features over the $T$ timesteps form $X = (x_1, x_2, \ldots, x_t, \ldots, x_T)^T$ as the input, where $x_t$ includes all the features for all the nodes at timestep $t$. Moreover, we used the percentage of inundated roads (determined based on INRIX traffic data) as the target variable and used $y_t^i$ to represent the flooding status of census tract $i$ at timestep $t$.

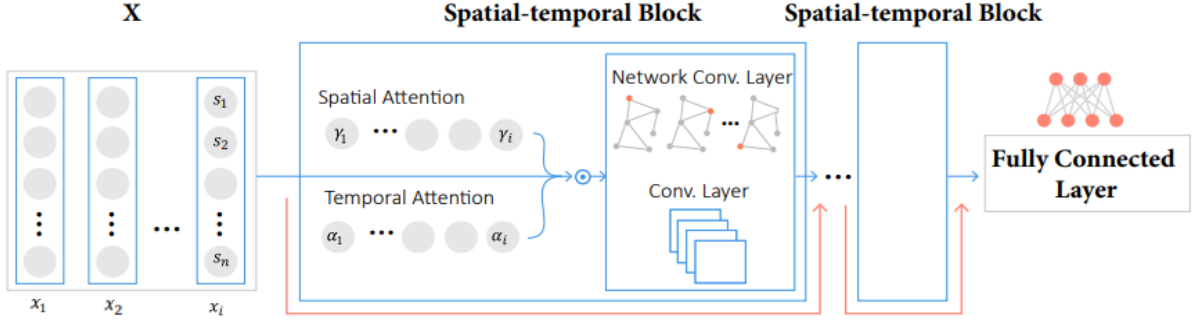

*Figure 4. Model architecture, including model input, spatial-temporal blocks, attention layers, and the fully connected layer at the end.*

As shown in Figure 4, the ASTGCN model consists of spatial-temporal (ST) blocks and a fully connected layer. Each ST block consists of a spatial attention module and a temporal attention module that is followed by a spatial-temporal convolution module on the graph model. The attention modules are included to capture the spatial and temporal correlation of the dynamic heterogeneous input features in the nowcasting flood status. These modules enable the network to adjust the weights of the features and determine the pieces of data upon which the model needs to rely more heavily to have generate predictions. The output is then fed into the spatial-temporal convolution module that captures the dependencies between different nodes based on the adjacency matrix and the time series of input features. The model includes $L$ ST blocks, where the input for $(l+1)^{th}$ block is:

$$X_h^l = (x_1, x_2, x_3, \ldots, x_{T_l}) \in \mathbb{R}^{N \times C_l \times \tau_l} \quad (1)$$

where $C_l$ denotes features of the input data in the $(l+1)^{th}$ layer, $\tau_l$ denotes the length of the temporal dimension in the $l^{th}$ layer, which for $l = 1$, equals $T$. The spatial attention is then determined as follows:

$$SAtt = P_s . \sigma(X^l W_1) W_2 (W_3 X^l)^T + b_s \quad (2)$$

where, $P_s$ and $b_s$ are $N \times N$ learnable parameters, and $W_{1_{C_l}}, W_{2_{C_l \times \tau_l}}$, and $W_{3_{C_l}}$ are also learnable parameters that are fed into sigmoid function $\sigma$ as the activation function. Similarly, the temporal attention module captures the strength of information between two timesteps $i$ and $j$. After processing at the attention modules, the data becomes more valuable for the convolution layer as it extracts and captures both dynamic spatial and temporal dependencies. The data is then fed into the spatial-temporal convolution module, which also has spatial and temporal dimensions. For applying convolution of the network structure, Guo et al. (2019) [80] used the spectral graph theory, and for each timestep, graph convolutions operate on the graph to extract correlation in the spatial dimension based on the developed adjacency matrix. Given $D$ as the degree matrix and $A$ as the adjacency matrix, Laplacian matrix ($L$) is defined as follows:

$$L = D - A \tag{3}$$

The normalized form of the Laplacian matrix is used to apply convolution on the graph as follows:

$$g_\theta *_G x = g_\theta(L)x = g_\theta(U\Lambda U^T)x = Ug_\theta(\Lambda)U^T x \tag{4}$$

where, $*_G$ operates a convolution on the graph $G$ given the signal $x$. Guo et al. (2019) [80] adopt a Chbyshev polynomial to approximate the eigenvalue decomposition on the Laplacian matrix and get the neighborhood of 0 to $k-1$-order of each node by $g_\theta$ as follows:

$$g_\theta *_G x = \sum_{k=0}^{K-1} \theta_k T_k(\tilde{L})x \tag{5}$$

where, $\theta$ consist of $K$ polynomial coefficients, $T_k(x) = 2xT_{k-1}(x) - T_{k-2}(x)$ and $\tilde{L}$ is determined as follows:

$$\tilde{L} = \frac{2}{\lambda_{max}} L - I_N \tag{6}$$

And $\lambda_{max}$ is the maximum eigenvalue of the Laplacian matrix. The Hadamard product of $T_k(\tilde{L})$ and $SAtt'$ is used in the approximation to include the effect of the spatial attention. Doing so, we can perform required number of filters for each node at each timestep and ensure that the neighboring information has been captured in the spatial dimension. Next, we use the similar standard temporal convolution to update the information based on the past timesteps; for the $l^{th}$ layer, we have:

$$X^l = ReLU\left(\Phi * \left(ReLU(g_\theta *_G X^{l-1})\right)\right) \tag{7}$$

where $*$ represents standard convolution, $\Phi$ parameters of temporal kernel, and $ReLU$ is the rectified linear unit activation function. The model in this study includes three appended ST blocks that are stacked to a fully connected layer that uses a softmax activation function for classifying the dependent variable, flood status.

Model Evaluation

We employed various classification metrics which can capture the performance of the model on the imbalance data, to evaluate the model performance. Accuracy, precision, recall, and F1 score are used for the case that the target variable is a categorical variable capturing the status of flooding. Our target variable has three different classes, thus we employed macro precision, recall, F1 score, and accuracy as model evaluation parameters to highlight the performance of the model on the minority class as follows:

$$p_{macro} = \sum_{i=1}^{M} p_i \tag{8}$$

where $p_{macro}$ is the macro precision of the model, $p_i$ is the precision for class $i$ and $M$ is the number of classes; $p_i$ is calculated as follows:

$$p_i = \frac{TP_i}{TP_i + FP_i} \tag{9}$$

where $TP_i$ is the number of true positive cases for class $i$ and $FP_i$ is the number of false positive cases for class $i$. Similarly, macro recall is calculated as follows:

$$r_{macro} = \sum_{i=1}^{M} r_i \qquad (10)$$

where $r_{macro}$ is the macro recall of the model, and $r_i$ is the recall for class $i$. $r_i$ is calculated as follows:

$$p_i = \frac{TP_i}{TP_i + FN_i} \qquad (11)$$

where $FN_i$ is the number of false negative cases for class $i$. In addition, macro F1 score is calculated as follows:

$$F1_{macro} = \sum_{i=1}^{M} f1_i \qquad (12)$$

where $F1_{macro}$ is the macro F1 score of the model, and $f1_i$ is the F1 score for class $i$. $f1_i$ is calculated as follows:

$$f1_i = \frac{p_i \times r_i}{p_i + r_i} \qquad (13)$$

Finally, we used model accuracy as the follows:

$$accuracy = \frac{\sum_{i=1}^{M}(TP_i + TN_i)}{\sum_{i=1}^{M}(TP_i + TN_i + FP_i + FN_i)} \qquad (14)$$

**Results**

Study Context

As one of the most flood-prone areas in the United States, Harris County has experienced several devastating floods since the latter half of the twentieth century. Notably, Hurricane Harvey, as a Category 4 hurricane, made landfall in Texas on August 25, 2017. Hurricane Harvey led to a catastrophic flood that necessitated 100,000 rescue requests in the week following its landfall in Harris County, as well as damage to 80,000 structures [74]. Contained within Harris County's 1,777 square miles are 22 primary watersheds. Detailed information regarding individual watersheds can be found at the Harris County Flood Control District website [81]. Each watershed has independent flooding management issues. Some of them merge and drain into one of the major creeks or bayous, but ultimately, all stormwater drains into Galveston Bay. We defined our study timeline from August 25, 2017, to September 2, 2017, and we collected the sets of data required for the flood nowcasting model for 787 census tracts in Harris County.

Implementation Details

In this study, we used data from August 25 to August 30, 2017, as our training set, and data from August 31, 2017, to September 3, 2017, as our test dataset. We used 30-minute intervals to give us 288 timesteps for training the data and 192 timesteps for testing the model. We split the data in a way that both training and testing sets capture portions of flood propagation and recession due to Hurricane Harvey. Accordingly, all the dynamic features were extracted and underwent data preprocessing required for feeding into the model. In cases that the dynamic features were not available in the time units of the study, linear interpolation and extrapolation were used to extract the required values for the missing timesteps. We categorized flooding statuses into three classes: in each timestep, census tracts with fewer than 1% of roads flooded are considered as "no flood," census tracts with 1%–10% of roads flooded are considered as "moderate flood, and census tracts with more than 10% roads flooded are considered as "severe flood." In this case, the model solves a classification problem in which the objective is to minimize the misclassified samples. We performed hyperparameter tuning by focusing on the learning rate and dropout rate to select the model with the best performance.

Model Performance and Comparison

Along with model implementation and to better evaluate the model performance, we used different state-of-the-art models against which ton compare the performance of the ASTGCN model. Moreover, we examined the extent to which the integration of human-sensed data can improve the performance of a model that relies solely on physics-based data for flood nowcasting. To this end, we ran four different experiments. First, we ran the model on the attention-based spatial-temporal graph convolution network model fed by physics-based data (model 1). Next, we employed the same ASTGCN model and employed both physics-based and human-sensed features as input (model 2). To assess the impact of the attention mechanism on the model performance, we used a relatively similar spatial-temporal graph convolutional network (STGCN) model (model 3) adopted from Yu et al. (2018) [82]. Finally, we used a long-short term memory (LSTM) model (model 4) as the baseline for model performance comparison. Table 2 shows the performance of the models in terms of precision, recall, F1 score, and model accuracy. Comparing the performance of graph-based models (models 1, 2, and 3) with the LSTM model, we can see that the graph-based models show significantly better performance in terms of precision, recall, and F1 score, while all the models have proper accuracy. The poor performance of the LSTM model in macro precision, recall, and F1 score shows that the model is unable to classify minority classes (i.e., flooded areas), which indicates that the model cannot provide insight for flood nowcasting. Comparing the performance of graph-based models, the STGCN model demonstrates highest recall and accuracy. However, the precision is 9.28% lower than the model with the highest precision, model 2, which uses physics-based and human-sensed input. This considerable difference is also reflected in the F1 score. The implication is that model 3 properly captures flooded cases (high recall), which is particularly valuable for flood nowcasting since it ensures the majority of the flooded areas are captured; however, the downside is that it erroneously captures many non-flooded cases.

*Table 2. Evaluation metrics for performance comparison of different models*

| Criteria/Model | ASTGCN-I[*] (model 1) | ASTGCN-II[**] (model 2) | STGCN (model 3) | LSTM (model 4) |
| --- | --- | --- | --- | --- |
| Precision | 0.785 | **0.808** | 0.733 | 0.416 |
| Recall | 0.824 | 0.891 | **0.906** | 0.413 |
| F1 Score | 0.802 | **0.842** | 0.819 | 0.414 |
| Accuracy | 0.975 | 0.979 | **0.999** | 0.981 |

[*]ASTGCN with physic-based features
[**]ASTGCN with physic-based and human-sensed features

Finally, the comparison of model 1 and model 2 reveals valuable insights for flood nowcasting and risk prediction. As shown in Table 2, model 2 over-perform model 1 in major evaluation metrics, including precision, recall, and F1 score. Particularly, model 2 yields 2.92% higher precision, 8.13% higher recall, and a 4.99% higher F1 score. Therefore, it can be seen that the use of human-sensed features as the supplement to physics-based input for flood nowcasting in the graph-based model significantly improves the predictive performance of the model. This finding shows the benefit of using heterogeneous community data and integrating different dynamic features for flood nowcasting. It reinforces the need for developing pipelines for collecting, preprocessing and integrating human-sensed data that becomes available during a flood event to improve awareness.

Figure 5 shows an instance of the prediction performance for model 2. As can be seen in boxes (I), (II), and (III), the model performed well in the case of the clusters of flooded areas, although in some cases (box (II)), there are misclassified regions. These region errors might indicate the impact of capturing the spatial dependency on the predictive performance that enables the model to identify the inundation hot spots and aid decision-makers to detect regions that need to be prioritized for emergency response in near future. On the other hand, as we can see in the red circles in Figure 5 (b), particular areas that are not in the flooded clusters have been classified incorrectly. This result might indicate the need for more data, particularly human-sensed data, which can signal inundation of areas where flooding is difficult to detect by the co-location dependency.

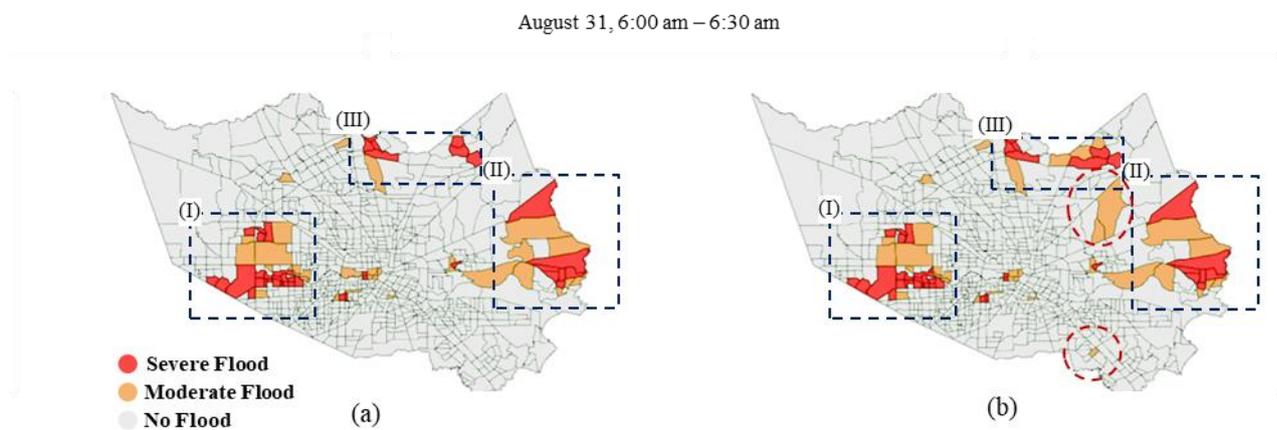

*Figure 5. An example of model overall predictive performance (August 31, 6:00 a.m.– 6:30 a.m.); (a) ground truth versus (b) model prediction*

Figure 6 also shows two cases of flood nowcasting performance by model 2, which shows significant differences in predictive performance. As shown in Figure 6 (a), the model has performed properly in identifying the majority of the flooded area; however, in considerable misclassified areas are evident in Figure 6 (b). Considering that Figure 6 (a) shows a timestep close to the start time of the test set (timestep 2), while Figure 6 (b) is a timestep that captures the third day in the test set (timestep 136), it might be inferred that the model performance decays as the time passes, which can be addressed by updating the model during the flood event.

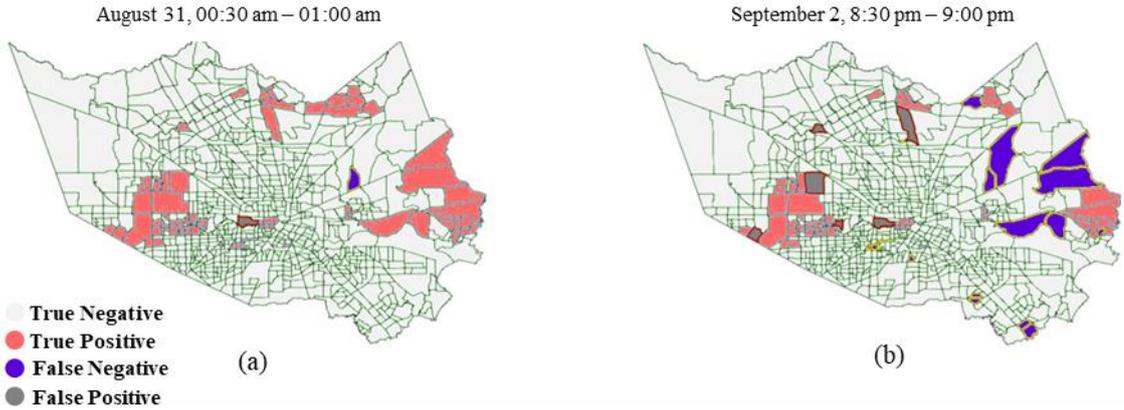

*Figure 6. An example of model flood nowcasting performance; (a) a proper prediction (August 31, 00:30 a.m.–01:00 a.m.) versus (b) model prediction with more misclassification (September 2, 8:30 p.m.–9:00 p.m.)*

**Concluding Remarks**

A crucial step for effective and timely disaster response and recovery is situational awareness, how the situation is evolving, and how community actors and residents respond to the evolving situation [83]. In this regard, flood nowcasting plays a pivotal role in enhancing situational awareness by providing a realistic prediction of the areas at risk of flood inundation in near future. In this study, we developed an attention-based spatial-temporal graph convolution network (ASTGCN) model for urban flood forecasting. The model employs both physics-based and human-sensed features, as well as static features that capture spatial dependency in terms of flood propagation. In the ASTGCN model, the attention mechanism enables automatically updating the importance of spatial and temporal dependencies for flood nowcasting, and the spatial and temporal convolutions extract the local dependencies in the model. We demonstrated the application of the model and compared its performance in the context of flooding following Hurricane Harvey in Harris County, Texas, in August 2017. The results indicate that, in general, the graph-based structure significantly improves prediction of flooded areas. For example, the model performs significantly better than the conventional long-term short-term models in terms of precision and recall, which are metrics of interest in prediction tasks using an imbalanced data set. Moreover, the attention mechanism improves the model recall and helps to capture the majority of flooded areas. The results also indicate that the ASTGCN model performs significantly better if it employs heterogeneous human-sensed data as a supplement of the physics-based data that traditionally used by hydraulic and hydrologic models. This finding is particularly significant since it demonstrates the promise of developing data pipelines for data fusion using physics-based data collected by flood gauges and sensors and data that is either generated by residents or captures the digital trace of residents' activity.

The main contributions of this study are twofold: first, we proposed and tested a novel graph deep-learning model for urban flood nowcasting. Second, the study showed the value of leveraging human-sensed data to complement physical flood sensor data for observing flood status across a region to improve flood nowcasting. Through these contributions, this study advances the body of knowledge related smart flood resilience. The advances in a structured deep-learning model provide opportunities for employing model architectures that extract information from spatial and temporal dependencies [2] and modules that extract information by putting more attention on the varying spatial and temporal features [42]. Moreover, the increasing availability of heterogeneous human activity data in near real-time calls for pipelines that leverage the information embedded in such data that can provide signals for urban flooding. The novel deep learning-modeling approaches and the availability of human-sensed data advance smart flood resilience by providing tools and pipelines that help people better respond and react to floods through enhanced predictive flood exposure and risk mapping before and during floods. This study, in particular, demonstrates the promise of integrating physics-based and human-sensed data into a graph-based deep-learning model that captures spatial and temporal dependencies for flood nowcasting. Also, this study showed the promise of data-driven models to complement physics-based H&H models for predictive flood monitoring and situational awareness. Future studies can focus on developing techniques to reduce the computational demand of the existing models to make the use of these models more feasible for flood nowcasting once more data streams are fed into the model. Moreover, further studies can generalize the approach demonstrated in this paper by testing the model on other flood cases and utilizing other types of physics-based and human-sensed features as inputs. One limitation of this study is that the model was tested in a single event and region, as the data used in this study was not available for historical events. As various physical sensor and human-sensed data become more available in future events, however, the model could be employed and tested in other events and contexts.

**Acknowledgements**
The authors would like to acknowledge funding support from the National Science Foundation project CRISP 2.0 Type 2 #1832662: Anatomy of Coupled Human-Infrastructure Systems Resilience to Urban Flooding: Integrated Assessment of Social, Institutional, and Physical Networks and the funding support from the X-Grant program (Presidential Excellence Fund) from Texas A&M University. Any opinions, findings, and conclusion or recommendations expressed in this research are those of the authors and do not necessarily reflect the view of the funding agencies.

**Competing Interests**
The authors declare no competing interests.

**Data Availability**
All data were collected through a CCPA- and GDPR-compliant framework and utilized for research purposes. The data that support the findings of this study are available from Mapbox and INRIX, but restrictions apply to the availability of these data, which were used under license for the current study. The data can be accessed upon request by the data providers. Other data we use in this study are all publicly available.

**Code Availability**
The code that supports the findings of this study is available from the corresponding author upon request.